\documentclass{article}

\usepackage{microtype}
\usepackage{graphicx}
\usepackage{subcaption}
\usepackage{booktabs} % for professional tables
\usepackage{multirow}
\usepackage{multicol}
\usepackage{xcolor}
\usepackage{colortbl}
\definecolor{citecolor}{RGB}{34,139,34}
\definecolor{fengblue}{RGB}{180, 208, 235}
\definecolor{fenggreen}{RGB}{197, 224, 179}
\definecolor{fengorange}{RGB}{251, 229, 213}
\definecolor{demphcolor}{gray}{.5}
\definecolor{Graylight}{gray}{0.9}

\usepackage{tikz}

\newcommand{\dotfilled}[1]{%
  \tikz[baseline=-0.6ex]\fill[#1] (0,0) circle (2.0pt);%
}

\newcommand{\dotopen}[1]{%
  \tikz[baseline=-0.6ex]\draw[#1, line width=0.6pt] (0,0) circle (2.0pt);%
}

\newlength\savewidth\newcommand\shline{\noalign{\global\savewidth\arrayrulewidth
  \global\arrayrulewidth 1pt}\hline\noalign{\global\arrayrulewidth\savewidth}}
\newcommand{\tablestyle}[2]{\setlength{\tabcolsep}{#1}\renewcommand{\arraystretch}{#2}\centering\footnotesize}
\renewcommand{\paragraph}[1]{\vspace{1.25mm}\noindent\textbf{#1}}

\usepackage{pifont}
\newcommand{\xmark}{\ding{55}}
\newcommand{\cmark}{\ding{51}}
\newcommand{\bxmark}{\textcolor{orange!40}{\ding{55}}}
\newcommand{\bcmark}{\textcolor{gray}{\ding{51}}}

\usepackage{hyperref}

\usepackage[preprint]{icml2026}

\usepackage{amsmath}
\usepackage{amssymb}
\usepackage{mathtools}
\usepackage{amsthm}

\usepackage[capitalize,noabbrev]{cleveref}

\theoremstyle{plain}

\theoremstyle{definition}

\theoremstyle{remark}

\usepackage[textsize=tiny]{todonotes}

\icmltitlerunning{ViT-5: Vision Transformers for The Mid-2020s}

\begin{document}

\twocolumn[
  \icmltitle{ViT-5: Vision Transformers for The Mid-2020s}



  \icmlsetsymbol{equal}{*}

  \begin{icmlauthorlist}
    \icmlauthor{Feng Wang}{jhu}
    \icmlauthor{Sucheng Ren}{jhu}
    \icmlauthor{Tiezheng Zhang}{jhu}
    \icmlauthor{Predrag Neskovic}{jhu}
    \icmlauthor{Anand Bhattad}{jhu}
    \icmlauthor{Cihang Xie}{ucsc}
    \icmlauthor{Alan Yuille}{jhu}
  \end{icmlauthorlist}

  \icmlaffiliation{jhu}{Johns Hopkins University}
  \icmlaffiliation{ucsc}{UC Santa Cruz}

  \icmlcorrespondingauthor{Feng Wang}{wangf3014@gmail.com}

  \icmlkeywords{Machine Learning, ICML}

  \vskip 0.3in
]



\printAffiliationsAndNotice{}  % no special notice (required even if empty)

\begin{abstract}
This work presents a systematic investigation into modernizing Vision Transformer backbones by leveraging architectural advancements from the past five years. While preserving the canonical Attention–FFN structure, we conduct a component-wise refinement involving normalization, activation functions, positional encoding, gating mechanisms, and learnable tokens. These updates form a new generation of Vision Transformers, which we call ViT-5. Extensive experiments demonstrate that ViT-5 consistently outperforms state-of-the-art plain Vision Transformers across both understanding and generation benchmarks. On ImageNet-1k classification, ViT-5-Base reaches 84.2\% top-1 accuracy under comparable compute, exceeding DeiT-III-Base at 83.8\%. ViT-5 also serves as a stronger backbone for generative modeling: when plugged into an SiT diffusion framework, it achieves 1.84 FID versus 2.06 with a vanilla ViT backbone. Beyond headline metrics, ViT-5 exhibits improved representation learning and favorable spatial reasoning behavior, and transfers reliably across tasks. With a design aligned with contemporary foundation-model practices, ViT-5 offers a simple drop-in upgrade over vanilla ViT for mid-2020s vision backbones. Code is available at \url{https://github.com/wangf3014/ViT-5}.
\end{abstract}

\section{Introduction}

Since its introduction at the end of 2020, the Vision Transformer~\cite{vit} (ViT) has substantially reshaped visual encoding paradigms. Its close architectural alignment with language transformers~\cite{transformer} has given rise to a wide range of successful vision–language systems for multimodal understanding and generation~\cite{clip,florence,coca,llava,qwen2.5vl,bagel,qwen-image,gpt4o}. Over roughly the past five years, language models have undergone a series of systematic refinements beyond the original Transformer design~\cite{bert,gpt,llama,qwen,mistral,falcon,gemma,phi1,deepseek,gptoss}. These include the adoption of more advanced activation functions and normalization layers, a transition from absolute to relative positional encodings, as well as improved attention normalization and gating mechanisms. Collectively, these structural evolutions have led to substantial gains in representational capacity and training stability, and have been a major driving force behind the rapid progress of modern foundation models.

Nonetheless, in contrast to the rapid and continuous refinement of language model architectures, the core design of ViTs has remained largely unchanged since their inception. For example, the state-of-the-art plain ViT backbone DeiT-III~\cite{deit3} introduces only minimal modification termed LayerScale~\cite{cait}, while otherwise preserving the original ViT architecture. Similarly, recent large-scale vision–language models such as SigLIP-2~\cite{siglip2} and Qwen3-VL~\cite{qwen3vl} continue to rely on essentially the same vanilla ViT design for visual encoding. This relative stagnation in architectural evolution raises an important question: \textit{is the representational potential of ViTs still under-optimized?} Given the substantial gains achieved by structural refinements in language models, a natural hypothesis is whether these advancements, spanning normalization, activation functions, positional encoding, and attention mechanisms, can be systematically transferred to vision models to unlock further performance and efficiency improvements.

In this work, we present a systematic study on optimizing the architectural design of ViTs. Our investigation centers around the original plain ViT formulation, deliberately preserving its fundamental Attention–FFN backbone, while focusing on identifying the most effective variants of individual architectural components. Rather than proposing a radical redesign, we aim to understand how modernized design choices can be incorporated into ViTs in a principled and modular manner. We explicitly analyze and empirically validate the impact of structural refinements that have emerged since the introduction of ViT such as LayerScale~\cite{cait}, Rotary Positional Embeddings (RoPE)~\cite{rope}, QK normalization~\cite{gemma3,qwen3}, and Register tokens~\cite{register}, many of which have become standard in either vision or language transformers. Through extensive experiments, we arrive at two key observations. \textbf{\textit{First, the current ViT architecture remains under-optimized}}, and refining its core components can consistently yield significant performance gains across diverse tasks. \textbf{\textit{Second, existing architectural refinements are not strictly orthogonal}}: naively combining all modern components does not necessarily lead to optimal performance, and effective modernization requires careful design.

\begin{figure}[t]
    \centering
    \includegraphics[width=\linewidth]{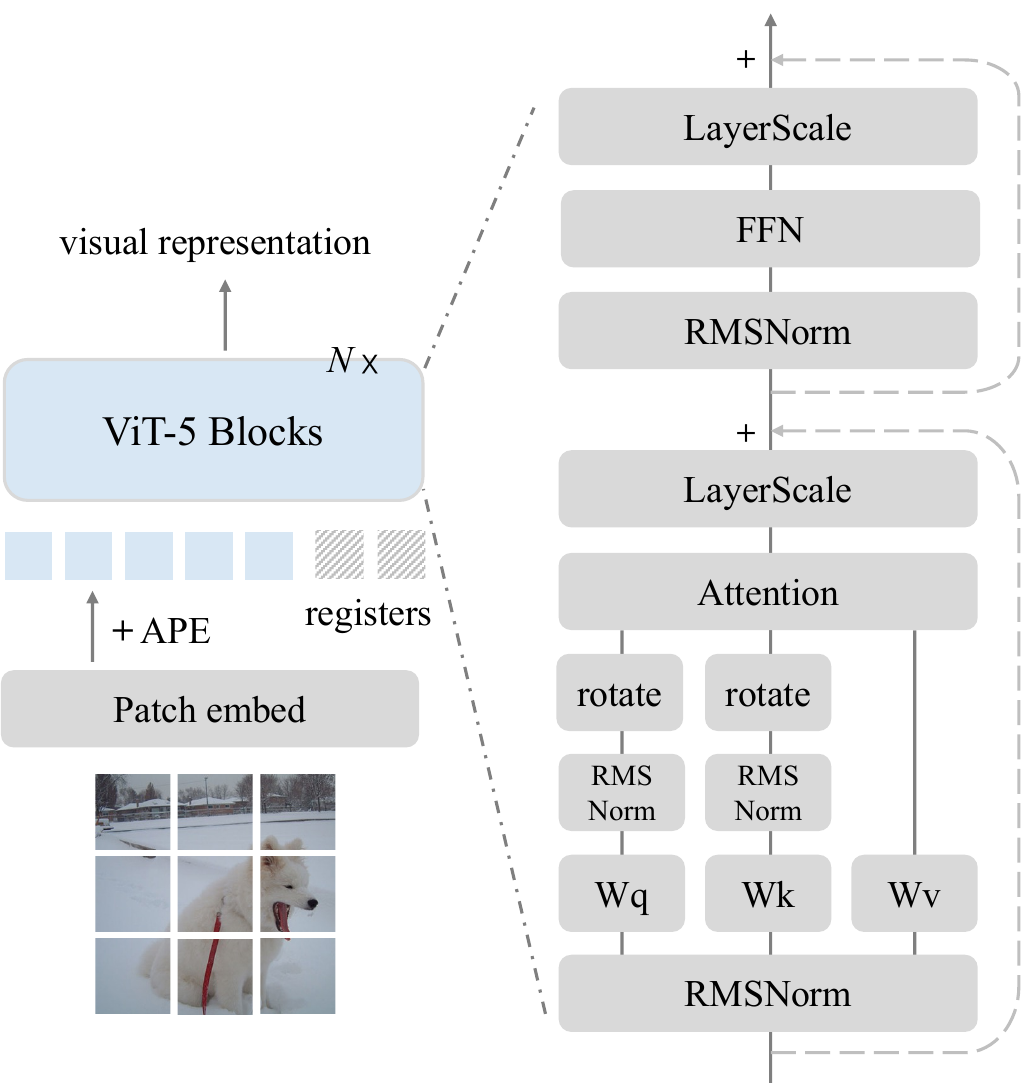}
    \caption{\textbf{\textit{Overview of ViT-5 architecture.}} We conduct in-depth analyses and modernize the ViT architecture by refining its components including activation scaling, normalization, positional embeddings, registers, bias terms, and \emph{etc}.}
    \label{fig:overview}
    % \vspace{-0.5cm}
\end{figure}

Motivated by these findings, we introduce a next-generation Vision Transformer architecture that comprehensively refines the original design. We term this new model \textbf{ViT-5}, which denotes the incorporation of the major architectural evolutions \textbf{\textit{over the past five years}}. We present an architectural overview of ViT-5 in Figure~\ref{fig:overview}. In detail, ViT-5 introduces a set of components that help stabilizing ViTs such as LayerScale, RMSNorm, and QK-Norm, as well as modules that enhance spatial reasoning such as RoPE and register tokens, while deliberately avoiding SwiGLU activations, which, despite their popularity in modern LLMs, can lead to over-gating issues (defined in Section~\ref{sec:gate}) in vision models. Our experiments across a range of vision tasks demonstrate the strong representational capacity and good generalizability of ViT-5. On standard ImageNet-1k classification, a base-sized ViT-5 achieves 84.2\% top-1 accuracy, outperforming the previous state-of-the-art plain ViT baseline, DeiT-III, which attains 83.8\%. For image generation, diffusion models built on ViT-5 achieve a FID of 1.84, significantly improving upon the SiT~\cite{sit} baseline’s 2.06 under nearly identical computational cost. Moreover, results on dense prediction tasks, together with quantitative evaluations, further highlight ViT-5’s advantages in spatial modeling and representation learning. 

This study improves the representational capacity of ViT backbones and narrows the architectural gap between vision models and modern language models. We hope ViT-5 can facilitate more efficient construction of multimodal systems and inspire the development of unified Transformer architectures that generalize seamlessly across modalities.

\section{Related Work}

\paragraph{Modern vision backbones.} The introduction of Vision Transformers marked a shift in visual representation learning from convolutional backbones~\cite{alexnet,lenet,vgg,resnet,densenet,efficientnet} to transformer-based architectures~\cite{vit}, offering a wide range of advances in visual pretraining~\cite{beit,mae,dinov2,dinov3,mocov3}, multimodal modeling~\cite{clip,align,coca,blip,siglip,sclip}, and image generation~\cite{dit,sit}. Following the ViT architecture, the DeiT series~\cite{deit,cait,deit3} introduce more efficient training strategies, enabling plain Vision Transformers to match the performance of contemporary CNN backbones such as ConvNeXt~\cite{convnext,convnextv2}, under comparable computational budgets. Beyond plain ViTs, a large body of work has explored hierarchical Vision Transformers~\cite{swin,swinv2,pvt,hiera,metaformer,cvt,focal-attn} and hybrid ViT–CNN architectures~\cite{coatnet,maxvit,levit}, which incorporate stronger inductive biases and often achieve superior performance on moderate-scale visual recognition benchmarks. This work does not aim to maximize absolute performance on specific benchmarks such as ImageNet~\cite{imagenet} or COCO~\cite{coco}. Instead, we focus on component-wise architectural optimization of plain Vision Transformers, with the goal of preserving their strong generalizability and scalability~\cite{vit22b,patch_scale} across model sizes and diverse visual tasks.

\paragraph{Architectural advancements for Transformers.} The Transformer~\cite{transformer} architecture has undergone a series of element-wise refinements, most prominently driven by the rapid evolution of large language models~\cite{llama,llama2,llama3,qwen,qwen2,qwen3,gemma,gemma2,gemma3,falcon,falcon3,deepseek,gptoss,phi1,phi4,mistral,mistral3,glm}. For example, early LLaMA~\cite{llama} models introduce rotary positional embeddings~\cite{rope} and RMS normalization, and replace the standard MLP with SwiGLU-based feed-forward networks~\cite{swiglu}. More recently, Gemma3~\cite{gemma3} further incorporates explicit normalization for queries and keys in self-attention, while Qwen3~\cite{qwen3} removes biases in the QKV projections to improve training stability and efficiency. Similar trends have also emerged in vision models. CaiT~\cite{cait} introduces LayerScale to stabilize the optimization of deep Vision Transformers, and subsequent work has shown that incorporating register tokens~\cite{register} can effectively mitigate activation artifacts in Transformers. Importantly, these refinements are centered around the original Transformer formulation, preserving the canonical attention–FFN inference structure without altering the overall model topology. How to effectively leverage this class of component-wise architectural improvements forms the primary focus of our study.

\section{Model}

\begin{table}[t]
    \centering
    \tablestyle{7pt}{1.2}
    \begin{tabular}{lccc}
    model & small & base & large \\\shline
    \textit{ViT-5 with post-normalization} & 82.18 & 84.15 & 84.82 \\\rowcolor{gray!15}
    \textit{ViT-5 with LayerScale} & 82.16 & 84.16 & 84.86 \\
    \end{tabular}
    \vspace{+0.1cm}
    \caption{\textbf{\textit{Replacing LayerScale by post-norm}} yields highly similar ImageNet top-1 accuracy across different model sizes.}
    \label{tab:postnorm}
    % \vspace{-0.6cm}
\end{table}

\subsection{Activation Scaling}

Among early refinements to ViTs, it is observed that introducing a learnable scaling factor on the output of each block can significantly improve training stability and performance for deep models~\cite{cait}. Formally, given an Attention or MLP block $\mathcal{F}(\cdot)$, it applies the following transformation:
\begin{equation}
    \mathbf{x}_{l+1} = \mathbf{x}_{l} + \mathcal{F}(\mathbf{x}_{l}) \odot \lambda,
\end{equation}
where $\lambda\in\mathbb{R}^d$ is a learnable scaling vector which is typically initialized to a small value (\emph{e.g.}, 10$^{-4}$). This mechanism is commonly referred to as LayerScale and has been used as a default component in many modern ViT architectures such as DINO v3~\cite{dinov3}.

While LayerScale has not been widely adopted in language models, we observe an intrinsic connection between LayerScale and post-normalization, which has been recognized as an important technique for stabilizing the training of deep LLMs. Formally, post-RMSNorm can be rewritten as
\begin{equation}
    \mathbf{x}_{l+1} = (\mathbf{x}_{l} + \mathcal{F}(\mathbf{x}_{l})) \odot \lambda_p / \text{Norm}
\end{equation}
where
\begin{equation}
    \text{Norm} = \text{RMS}(\mathbf{x}_{l} + \mathcal{F}(\mathbf{x}_{l}))
\end{equation}
and $\lambda_p\in\mathbb{R}^d$ is a scaling vector. We can find that LayerScale directly controls the scale of the block output, while post-normalization implicitly scales both the block output and residual. In our experiments, LayerScale and post-normalization lead to highly similar performance improvements (see Table~\ref{tab:postnorm}). Given that LayerScale offers greater flexibility and lower computational overhead, \underline{\textbf{\textit{we introduce LayerScale as a default component in ViT-5}}}. We hope that this observation on the functional relationship between LayerScale and post-normalization may also provide useful insights for future LLM architecture design.

\begin{table}[t]
    \centering
    \tablestyle{7pt}{1.1}
    \begin{tabular}{cccc}
    LayerScale & SwiGLU MLP & acc. (\%) & FID ($\downarrow$) \\\shline
    \xmark & \xmark & 83.86 & 15.80 \\
    \xmark & \cmark & 83.94 & 15.48 \\
    \cmark & \cmark & 83.70 & 16.22 \\\rowcolor{gray!15}
    \cmark & \xmark & \bf 84.16 & \bf 14.57 \\
    \end{tabular}
    \caption{\textbf{\textit{The over-gating issue emerges}} when combining LayerScale and SwiGLU MLP. We report ImageNet-1k top-1 accuracy for a ViT-B and ImageNet-256 FID for a ViT-XL. The default setup of ViT-5 is highlighted. The best results are bolded.}
    \label{tab:swiglu}
    % \vspace{-0.5cm}
\end{table}

\subsection{Normalization}
Since influential language models such as LLaMA~\cite{llama}, PaLM~\cite{palm}, and Gopher~\cite{gopher}, the de facto standard in LLM architectures has largely shifted from Layer Normalization (LayerNorm) to Root Mean Square Normalization (RMSNorm). A common observation underlying this transition is that the re-scaling invariance of normalization layers dominates their practical effect. Removing the re-centering operation in LayerNorm does not degrade performance, and can even lead to slight improvements by reducing unnecessary shifting noise. We observe a similar phenomenon in ViTs. In our experiments, replacing LayerNorm with RMSNorm slightly reduces computational cost and yields a modest performance gain (\emph{e.g.}, +0.2\% top-1 accuracy on ImageNet for ViT-B). Motivated by these findings, \underline{\textbf{\textit{we use RMSNorm throughout ViT-5}}}, replacing all LayerNorm layers in the original architecture.

\subsection{Gated MLP}\label{sec:gate}
Similar to the transition from LayerNorm to RMSNorm, modern LLMs have widely utilized gated MLP architectures, in which the traditional GeLU activation is replaced by SwiGLU (Swish-Gated Linear Unit)~\cite{swiglu}. Nonetheless, in this study, we observe that combining SwiGLU MLP with LayerScale can lead to a notable performance degradation in ViTs (see Table~\ref{tab:swiglu}). We attribute this issue to the fact that both LayerScale and gated MLPs effectively perform channel-wise filtering, which increases the sparsity of intermediate representations; when used together, their combined effect can result in excessively sparse activations. From this perspective, LayerScale can be viewed as a form of static gating, and the performance drop observed when jointly using LayerScale and SwiGLU MLP can be understood as a case of \textbf{\textit{over-gating}}. 

Here we draw a preliminary conclusion: at least up to the ViT-XL scale (with a hidden dimension of 1152 and 449M parameters), the combination of LayerScale and gated MLPs should be avoided. For larger-scale models with higher hidden dimensions or substantially more parameters, the over-sparsity issue may be mitigated, but we leave a systematic investigation of this regime for future work. Accordingly, \underline{\textbf{\textit{ViT-5 uses the original MLP design with GeLU activation}}}.

\begin{figure}[t]
  \centering
  \hspace{0.7cm}
  \begin{subfigure}[b]{0.35\columnwidth}
    \centering
    \includegraphics[width=\textwidth]{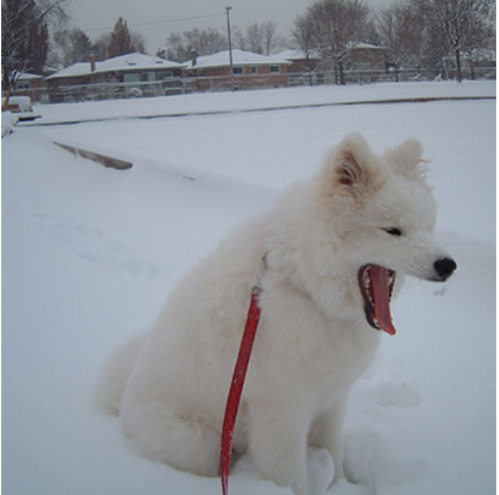}
    \caption{original}
    \label{fig:flip_0}
  \end{subfigure}
  \hfill
  \begin{subfigure}[b]{0.35\columnwidth}
    \centering
    \includegraphics[width=\textwidth]{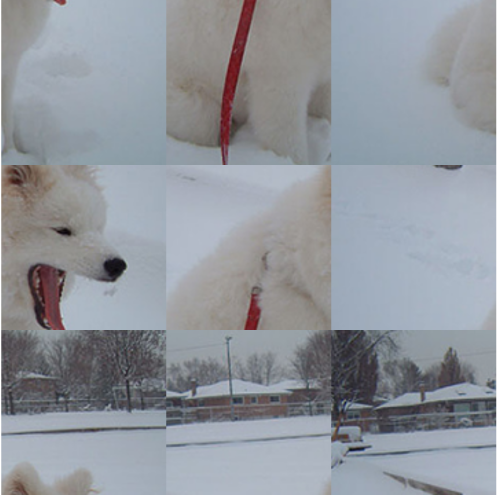}
    \caption{patch flipped}
    \label{fig:flip1}
  \end{subfigure}
  \hspace{0.7cm}
  \caption{\textbf{\textit{Undesired invariance when discarding APE}}. The two images are equivalent for ViTs with only RoPE as their positional embedding. Absolute position is needed for generic vision backbones, so ViT-5 takes both APE and RoPE as default components.}
  \label{fig:flip}
\end{figure}

\begin{table}[t]
    \centering
    \tablestyle{6pt}{1.2}
    \begin{tabular}{lccc}
    configuration & small & base & large \\\shline
    \textit{no register} & 82.04 & 84.02 & 84.61 \\
    \textit{vanilla registers} & 81.95 & 83.90 & 84.37 \\
    \textit{RoPE on registers, same freq. base} & 82.05 & 84.00 & 84.59 \\\rowcolor{gray!15}
    \textit{RoPE on registers, high freq. base} & \bf 82.16 & \bf 84.16 & \bf 84.86 \\
    \end{tabular}
    \caption{\textbf{\textit{Registers need relative positions.}} Vanilla registers leads to degraded performance. The issue is addressed by applying high-frequency RoPE to registers. Results denote IN1k test accuracy.}
    \label{tab:register}
    \vspace{-0.3cm}
\end{table}

\subsection{Positional Encoding}
Standard ViTs employ learnable absolute positional embeddings (APE), which have been shown to lack explicit relative positional modeling in complex visual reasoning tasks and to be inherently limited when handling dynamic input resolutions~\cite{qwen2vl,pixtral}. Following these findings, we extend rotary positional embeddings (RoPE) to the 2D setting and incorporate them into our models. Importantly, we do not discard absolute positional embeddings; instead, \underline{\textbf{\textit{ViT-5 jointly employs both APE and 2D RoPE}}}. The motivation is that using relative positional encoding alone can introduce undesirable invariances. For example, under a RoPE-only formulation, patch-level flips of an image become fully invariant, as illustrated in Figure~\ref{fig:flip} , where the two input images are treated as equivalent. While this behavior has limited impact on simple tasks such as image classification, it can pose potential limitations for a generic vision backbone, where absolute spatial cues may be critical for more complex visual reasoning.

\begin{figure}
    \centering
    \begin{subfigure}[b]{\columnwidth}
        \centering
        \vspace{-0.2cm}
        \includegraphics[width=\textwidth]{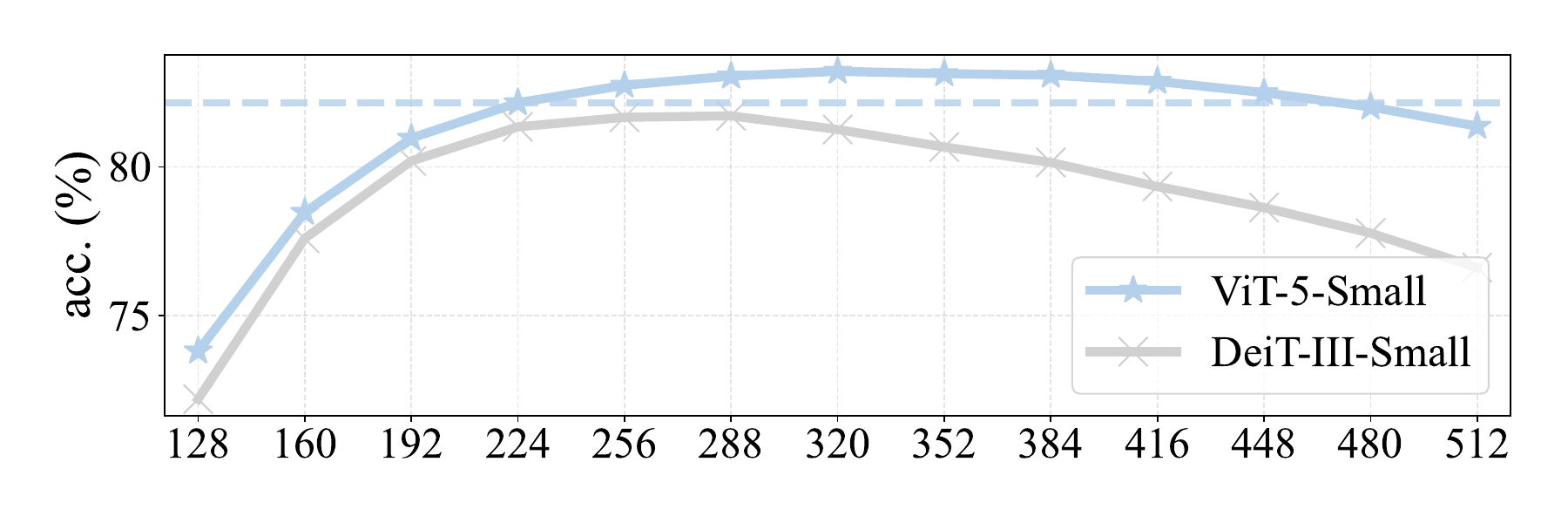}
  \end{subfigure}
  \begin{subfigure}[b]{\columnwidth}
        \centering
        \vspace{-0.3cm}
        \includegraphics[width=\textwidth]{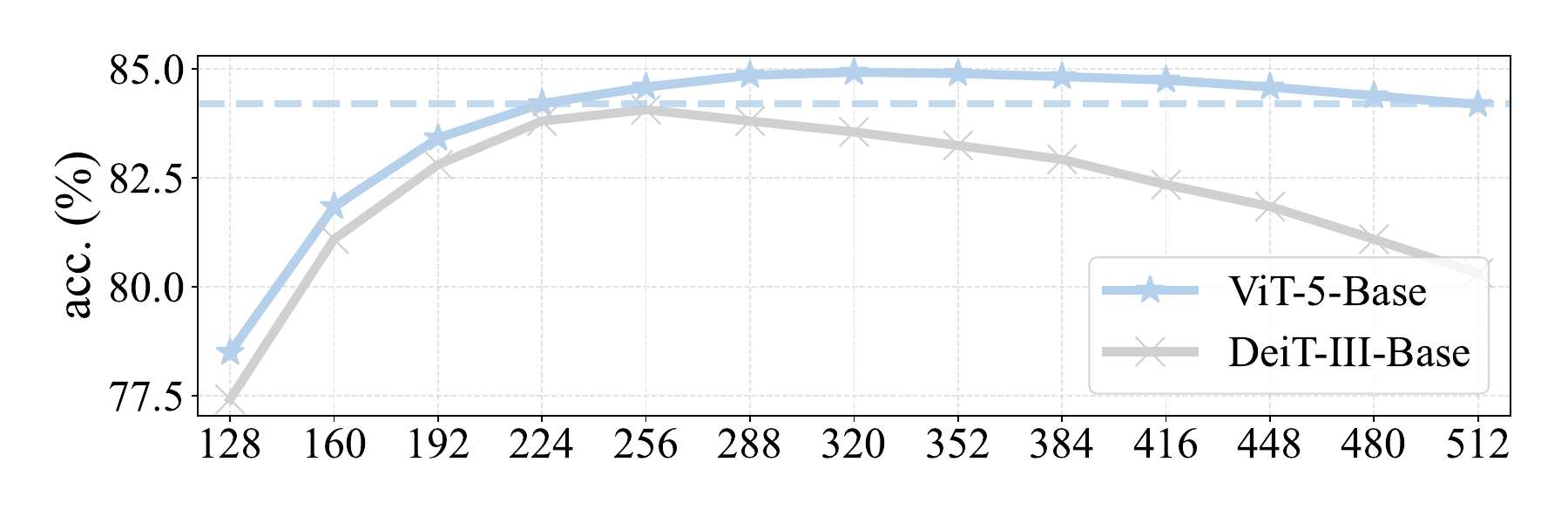}
  \end{subfigure}
  \begin{subfigure}[b]{\columnwidth}
        \centering
        \vspace{-0.3cm}
        \includegraphics[width=\textwidth]{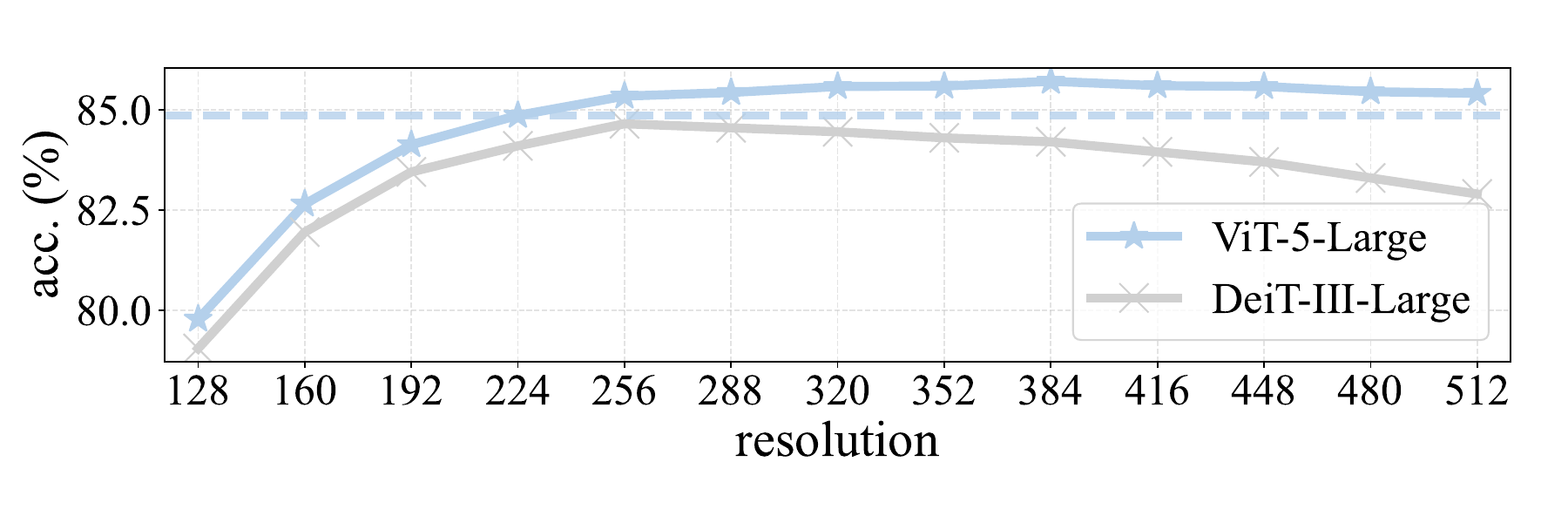}
        \vspace{-0.7cm}
  \end{subfigure}
  \caption{\textbf{\textit{Performance at dynamic resolutions.}} All models are trained at 224$^2$ and then tested at different input sizes.}
  \label{fig:resolution}
  % \vspace{-0.5cm}
\end{figure}

Figure~\ref{fig:resolution} compares the ability of ViT-5 and DeiT-III~\cite{deit3} to handle dynamic input resolutions. Both models are trained at a resolution of 224×224 and evaluated at different test resolutions without fine-tuning. The results show that DeiT-III, which relies solely on absolute positional embeddings, performs best only near the training resolution, and its accuracy degrades rapidly as the input size increases. In contrast, ViT-5 exhibits significantly stronger resolution robustness. Across a wide input range from 224 to 512, ViT-5 shows no noticeable performance degradation. Moreover, over a wide resolution span (\emph{e.g.}, from 128 to 384 for ViT-5-L), increasing the input size consistently leads to performance gains, indicating improved scalability and generalization to dynamic resolutions.

\begin{figure}[t]
    \centering
    \includegraphics[width=\linewidth]{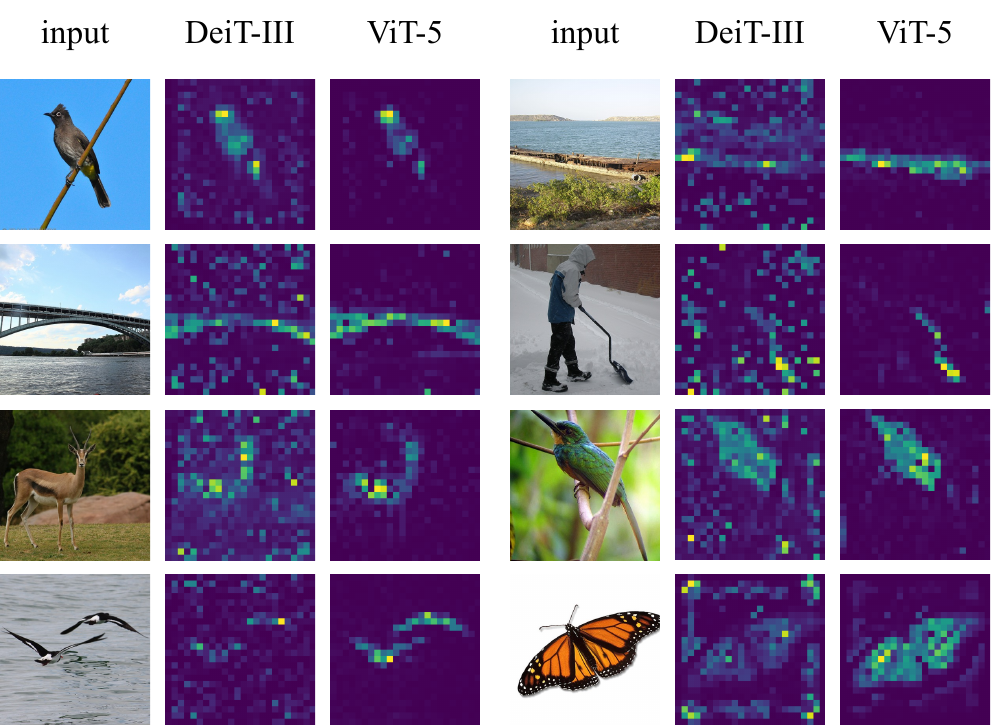}
    \caption{\textbf{\textit{Attention visualization}} of DeiT-III-L and ViT-5-L at 384$\times$384 resolution. ViT-5 exhibits improved spatial understanding, characterized by clearer and more accurate self-attention activations. This improvement primarily arises from the combined effects of relative positional embeddings and registers.}
    \label{fig:attn}
    \vspace{0.1cm}
\end{figure}

\subsection{Register Tokens}
\textit{``Vision Transformers Need Registers''} denotes a recent finding that the artifacts appear in ViTs can be effectively addressed by additional learnable tokens appended to input patch tokens~\cite{register}. In fact, the practical role of registers goes beyond their original motivation. These learnable tokens offer a flexible representation space, enabling extensions such as meta queries~\cite{metaquery} and 1D tokenization~\cite{titok} for images. To leverage these properties, \underline{\textbf{\textit{we introduce registers as a default component in ViT-5}}}.

We observe that in ViTs equipped with RoPE, register tokens should also be assigned relative positional embeddings. This is because when vector rotary operations are applied only to patch tokens while leaving register tokens unrotated, the registers naturally exhibit lower cosine similarity with patches that undergo smaller rotational angles. This imbalance distorts the attention distribution and implicitly introduces an undesired positional bias. To address this issue, \underline{\textbf{\textit{we equip register tokens with a separate 2D RoPE}}}, whose frequency base is significantly higher than that used for patch tokens. This design induces distinct rotational behaviors for registers and patch tokens across different channel dimensions, effectively decoupling their positional correlations and eliminating the undesired bias. Empirically, this modification stabilizes register–patch interactions and improves overall representation quality, with a detailed performance comparison shown in Table~\ref{tab:register}.

\begin{table}[t]
    \centering
    \tablestyle{5pt}{1.2}
    \begin{tabular}{lrrrrr}
         model & \#layers & dim & \#heads & \#registers & \#params \\
         \shline
         ViT-5-S & 12 & 384 & 6 & 4 & 22M \\
         ViT-5-B & 12 & 768 & 12 & 4 & 87M \\
         ViT-5-L & 24 & 1024 & 16 & 4 & 304M \\
         ViT-5-XL & 28 & 1152 & 16 & 4 & 449M \\
    \end{tabular}
    \caption{\textbf{\textit{Configuration of ViT-5 models.}} The setup aligns with the practice of existing ViT models~\cite{vit,dit}. We employ four learnable registers for all models.}
    \label{tab:config}
    \vspace{-0.2cm}
\end{table}

We also provide a qualitative comparison of attention activations between DeiT-III and ViT-5 in Figure~\ref{fig:attn}. Consistent with prior observations~\cite{register}, register tokens effectively suppress background artifacts in attention maps, enabling the class token to attend more accurately to semantically meaningful regions of the image. In ViT-5, we observe notably cleaner and more focused feature maps, which we attribute to the combined effects of register tokens and relative positional embeddings. Both components play an important role in enhancing the spatial modeling capability of Vision Transformers. Additional visualizations are provided in the Appendix.

\begin{figure}
    \centering
    \includegraphics[width=\columnwidth]{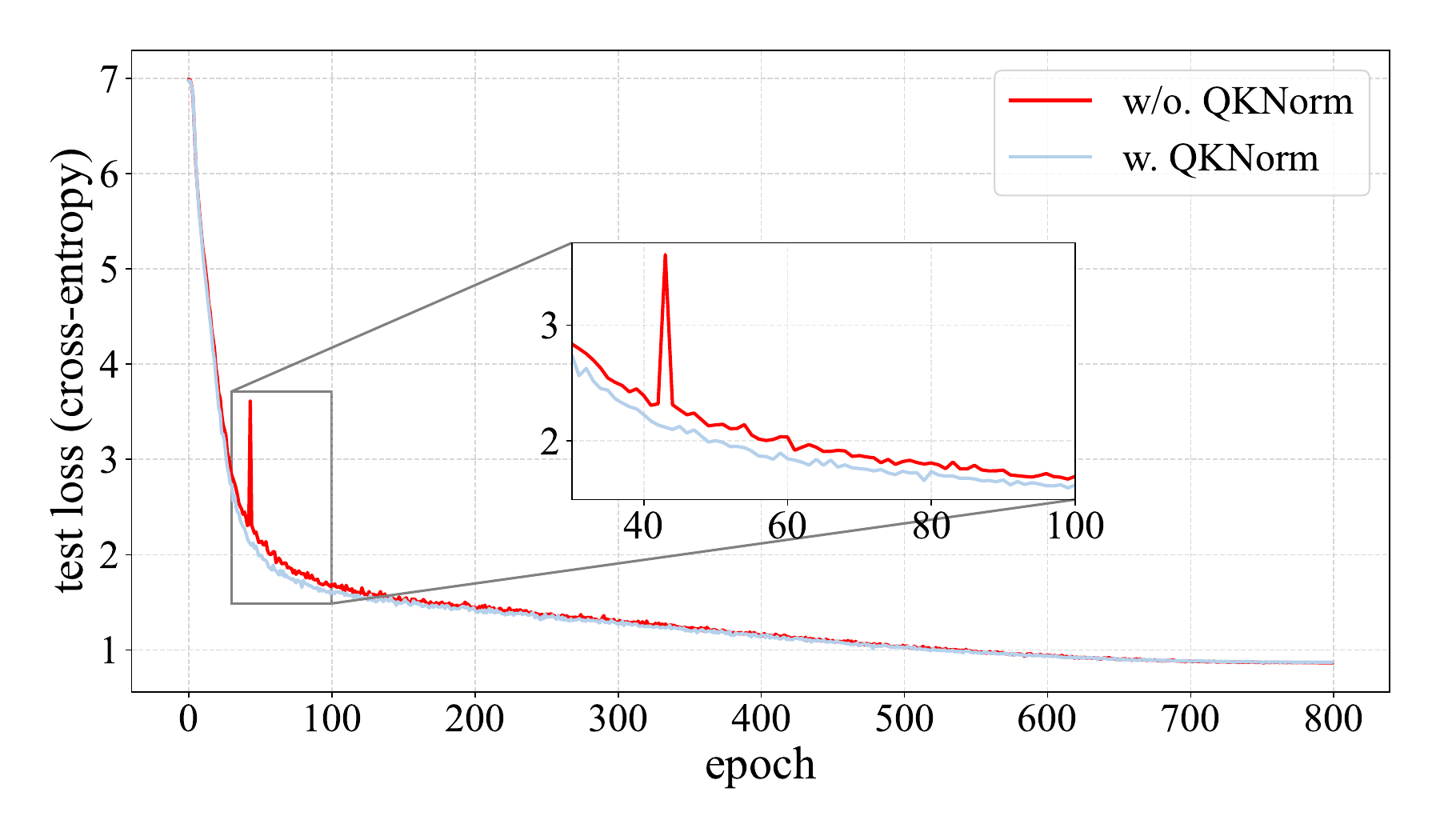}
    \vspace{-0.9cm}
    \caption{\textit{\textbf{QKNorm improves training stability.}} We show epoch-wise test loss of a ViT-5-Small with and without QKNorm, while the former converges smoothly and does not show spikes.}
    \label{fig:spike}
    % \vspace{-0.2cm}
\end{figure}

\subsection{QK-Normalization}

The latest LLMs such as Qwen3~\cite{qwen3} and Gemma3~\cite{gemma3} have begun to reform self-attention by applying additional normalization to the query and key. Formally, this QK-Normalization mechanism has
\begin{equation}
    Q' = \text{RMSNorm}(Q),\ K' = \text{RMSNorm}(K),
\end{equation}
\begin{equation}
    \text{Attention}(Q, K, V) = \text{Softmax}\left( Q'K'^T/\sqrt{d} \right) V.
\end{equation}
We find that QK-Normalization yields a modest performance improvement for ViTs, yet can significantly enhance training stability and reduce the occurrence of sharp loss spikes during optimization (a comparison is shown in Figure~\ref{fig:spike}). To leverage these benefits of robustness, \underline{\textbf{\textit{we introduce QK-Norm as a default component in ViT-5}}}.

\subsection{Bias Terms for QKV}
We utilize bias-free RMSNorm across the entire ViT-5 models, including all pre-norm and QK-Norm layers. This highlights the perspective that self-attention relies more on weighted projections than on additive biases. Consequently, \underline{\textbf{\textit{we remove bias terms in the QKV projection layers}}} to ensure structural consistency. This modification allows QK-Norm to operate more effectively and leads to noticeable performance improvements, as detailed in Section~\ref{sec:abl}.

\section{Experiments}
\label{sec:exp}

\begin{table}[]
    \centering
    \tablestyle{6pt}{1.2}
    \begin{tabular}{lrrrr}
         Model & Input & \#Params & FLOPs & Acc. (\%) \\
        \shline
        \dotfilled{fengblue} \textit{ConvNeXt-T}   & 224$^2$ & 29M  & 4.5G   & 82.1 \\
        \dotopen{orange} \textit{DeiT-III-S}   & 224$^2$ & 22M  & 4.6G   & 81.4 \\
        \rowcolor{gray!15}
        \dotfilled{orange} \textit{ViT-5-S}       & 224$^2$ & 22M  & 4.7G   & \bf 82.2 \\
        \hline
        \dotfilled{fengblue} \textit{ConvNeXt-S}   & 224$^2$ & 50M  & 8.7G   & 83.1 \\
        \dotfilled{fengblue} \textit{ConvNeXt-B}   & 224$^2$ & 89M  & 15.4G  & 83.8 \\
        \dotopen{orange}  \textit{DeiT-III-B}   & 224$^2$ & 87M  & 17.6G  & 83.8 \\
        \rowcolor{gray!15}
        \dotfilled{orange} \textit{ViT-5-B}       & 224$^2$ & 87M  & 17.9G  & \bf 84.2 \\
        \dotfilled{fengblue} \textit{ConvNeXt-B}   & 384$^2$ & 89M  & 45.0G  & 85.1 \\
        \dotopen{orange} \textit{DeiT-III-B}   & 384$^2$ & 87M  & 55.5G  & 85.0 \\
        \rowcolor{gray!15}
        \dotfilled{orange} \textit{ViT-5-B}       & 384$^2$ & 87M  & 55.9G  & \bf 85.4 \\
        \hline
        \dotfilled{fengblue} \textit{ConvNeXt-L}   & 224$^2$ & 198M & 34.4G  & 84.3 \\
        \dotopen{orange} \textit{DeiT-III-L}   & 224$^2$ & 304M & 61.6G  & 84.5 \\
        \rowcolor{gray!15}
        \dotfilled{orange} \textit{ViT-5-L}       & 224$^2$ & 304M & 62.8G  & \bf 84.9 \\
        % \hline
        \dotfilled{fengblue} \textit{ConvNeXt-L}   & 384$^2$ & 198M & 101.0G & 85.5 \\
        \dotopen{orange} \textit{DeiT-III-L}   & 384$^2$ & 305M & 191.2G & 85.4 \\
        \rowcolor{gray!15}
        \dotfilled{orange} \textit{ViT-5-L}       & 384$^2$ & 305M & 192.5G & \bf 86.0 \\
    \end{tabular}
    \caption{\textbf{\textit{ImageNet classification results.}} We compare ViT-5 with ConvNeXt~\cite{convnext} and DeiT-III~\cite{deit3}, which are representative for modern CNNs and ViTs. ViT-5 attains best performance in each group of parameter count and FLOPs.}
    \label{tab:cls}
    \vspace{-0.3cm}
\end{table}

\subsection{Image Classification}
Following prior practice~\cite{deit,deit3}, we train all models from scratch on ImageNet-1k~\cite{imagenet} by DeiT-III's recipe with small modifications. The detailed implementation is provided in the Appendix. The model configurations are summarized in Table~\ref{tab:config}. As shown in Table~\ref{tab:cls}, ViT-5 consistently achieves state-of-the-art performance under comparable parameter counts and computational budgets. Notably, its performance advantage scales smoothly with both model size and input resolution. In particular, ViT-5-L attains 86.0\% test accuracy with 384×384 inputs, significantly surpassing the previous state of the art for ViT (85.4\%) and CNN-based models (85.5\%). These results indicate that substantial and scalable gains can be achieved by systematically refining the components of ViTs. As ImageNet-1k remains the most widely used benchmark for evaluating vision backbones, the consistent improvements observed on this dataset provide strong evidence that ViT-5 offers enhanced representation learning capability and can serve as a competitive, generalized vision backbone to replace existing architectures.

\begin{figure*}[t]
    \centering
    \begin{subfigure}[b]{0.25\textwidth}
        \centering
        \includegraphics[width=\textwidth]{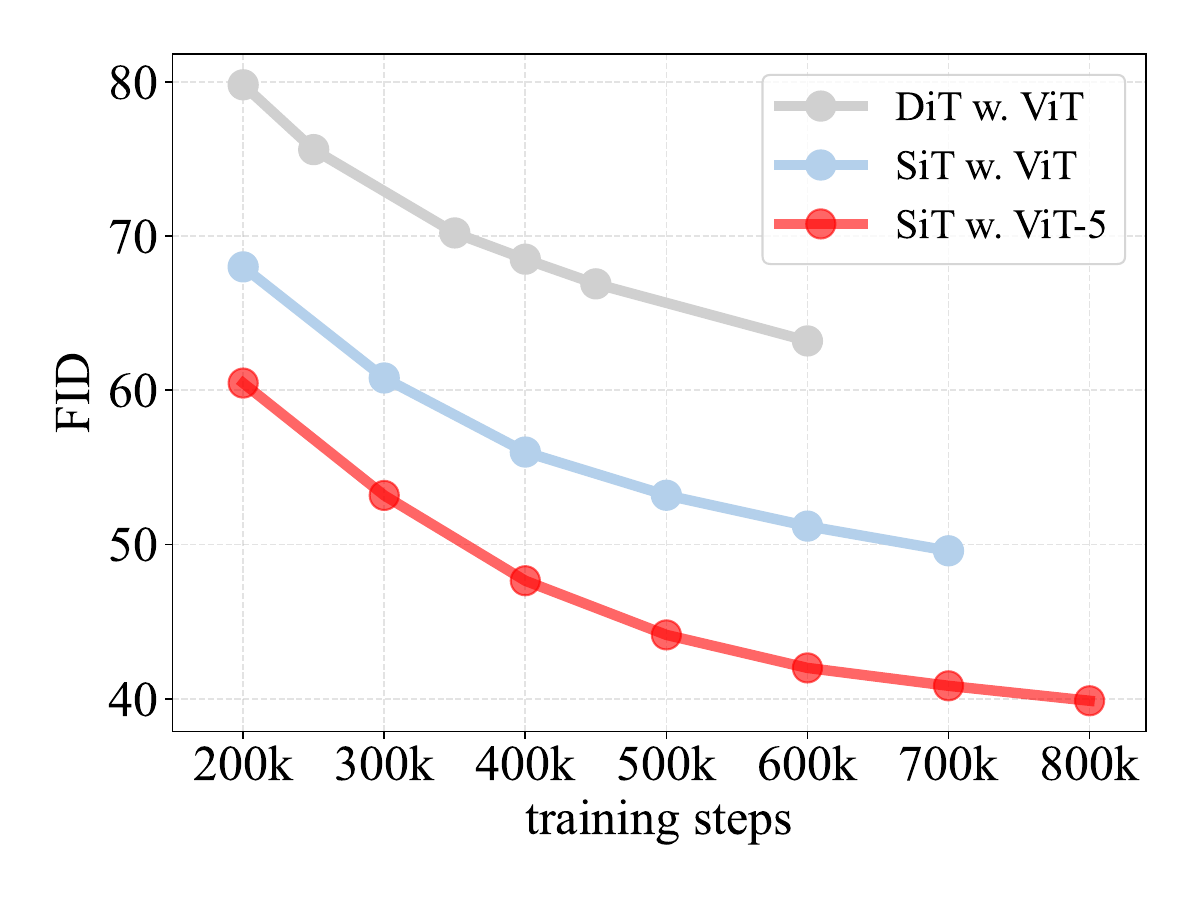}
        \label{fig:fid_small}
    \end{subfigure}
    \hspace{-10pt}
    \begin{subfigure}[b]{0.25\textwidth}
        \centering
        \includegraphics[width=\textwidth]{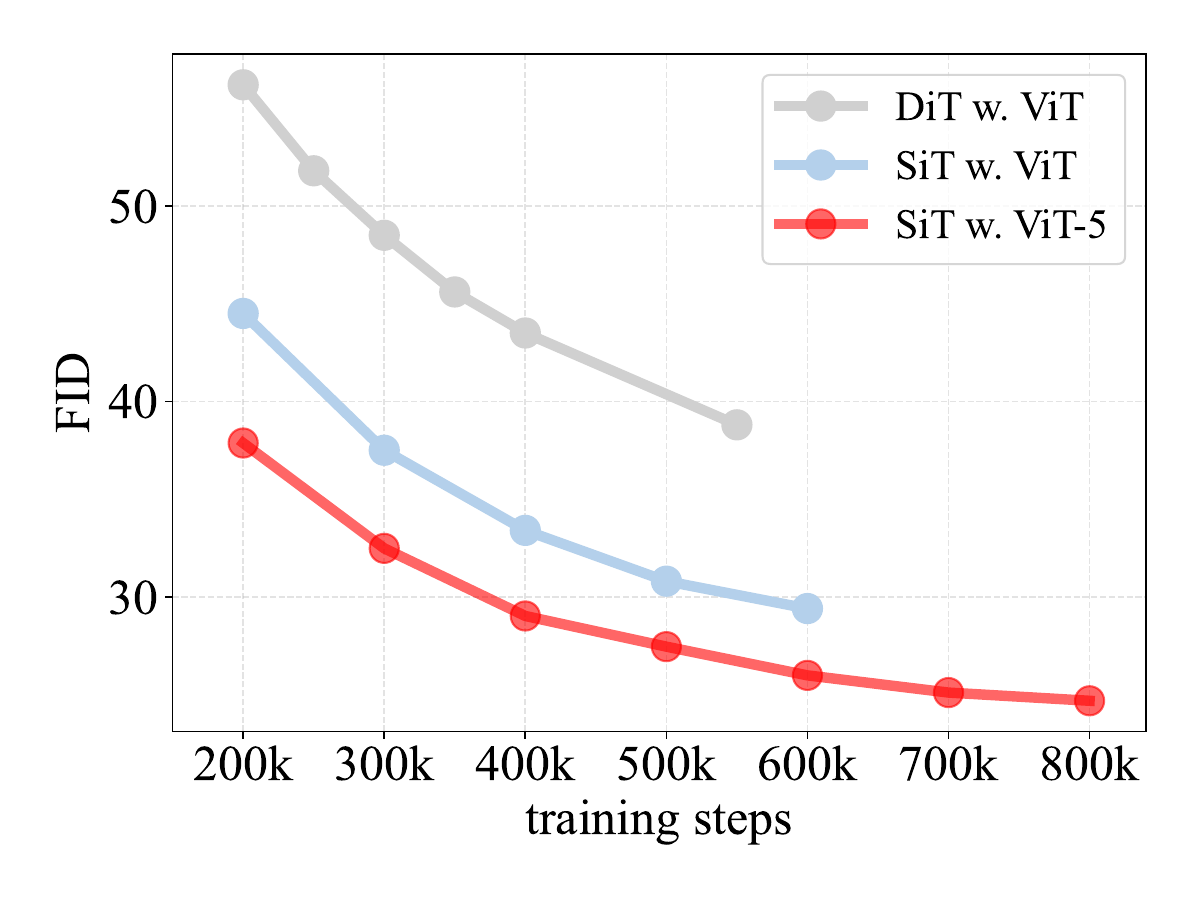}
        \label{fig:fid_base}
    \end{subfigure}
    \hspace{-10pt}
    \begin{subfigure}[b]{0.25\textwidth}
        \centering
        \includegraphics[width=\textwidth]{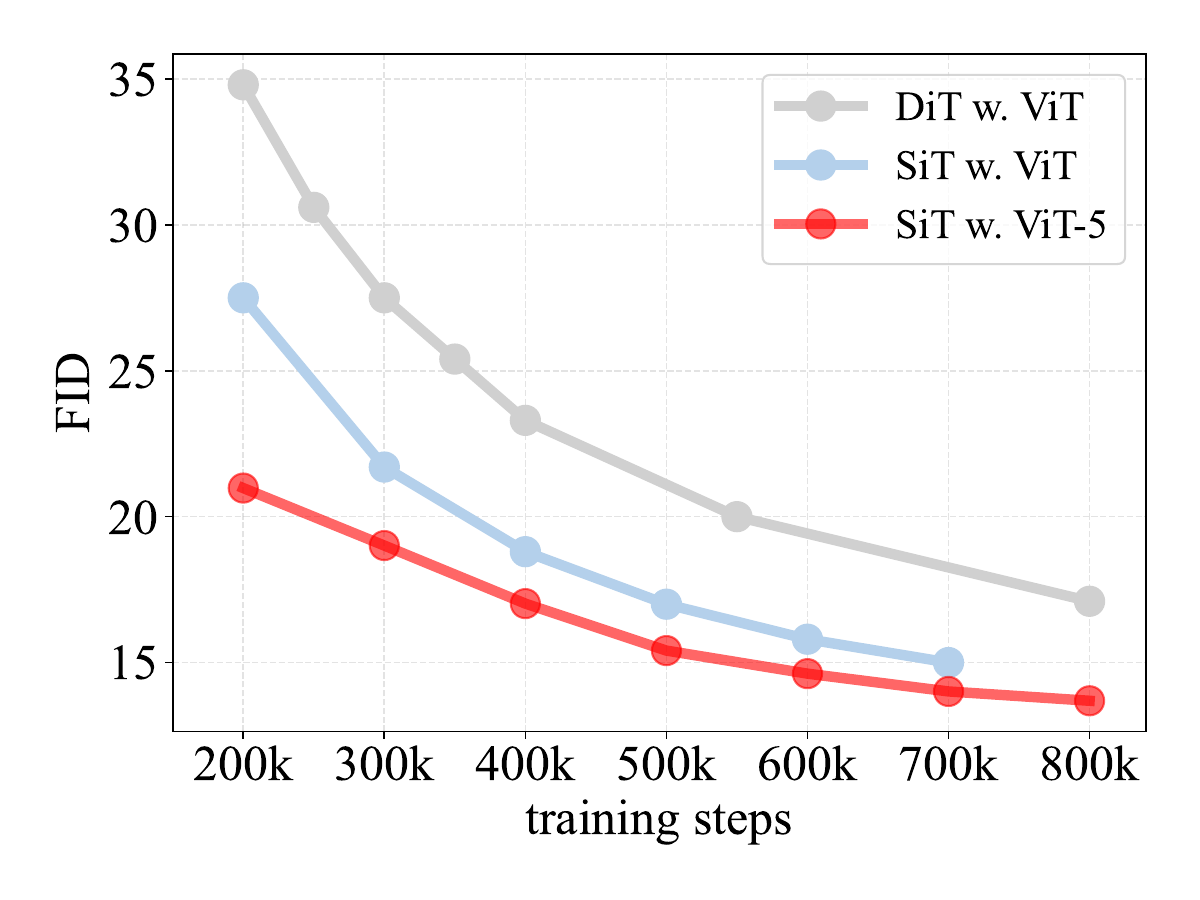}
        \label{fig:fid_large}
    \end{subfigure}
    \hspace{-10pt}
    \begin{subfigure}[b]{0.25\textwidth}
        \centering
        \includegraphics[width=\textwidth]{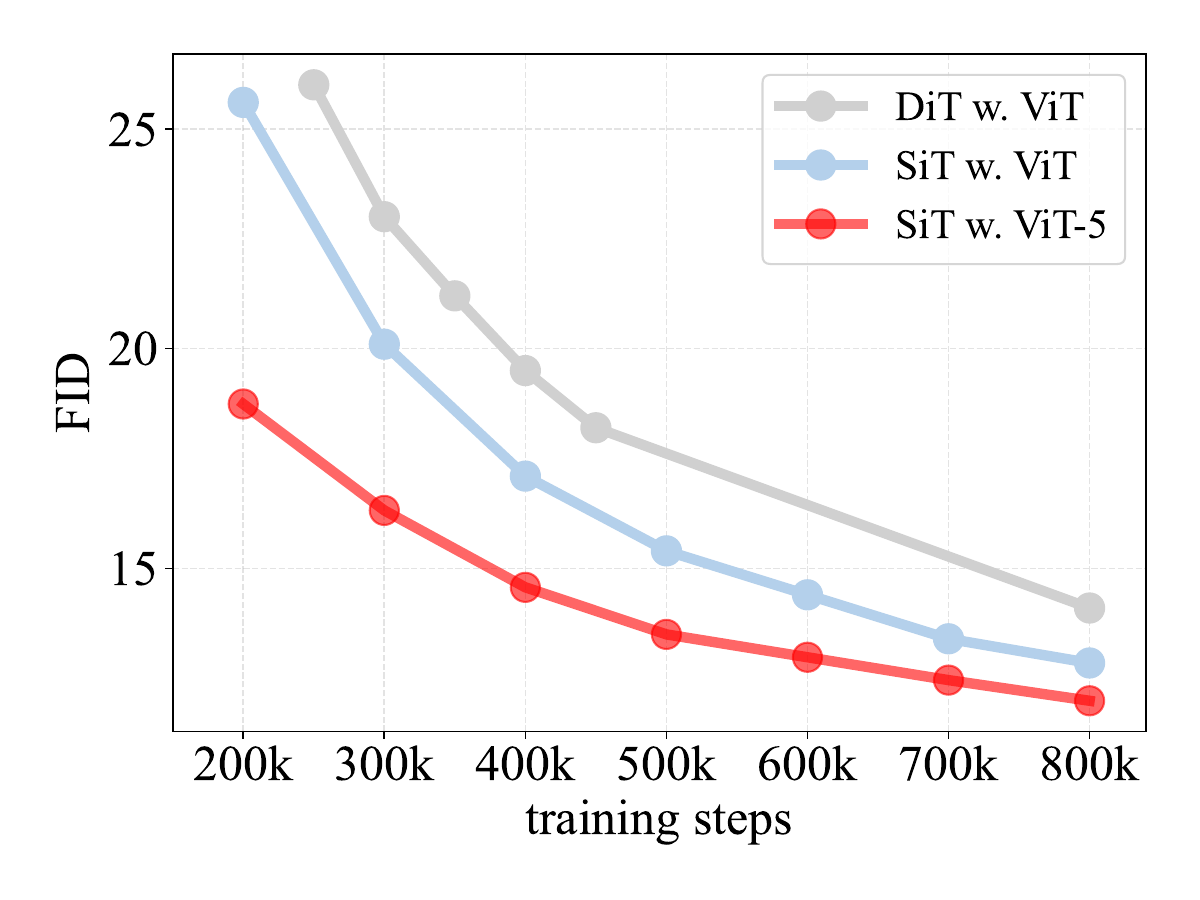}
        \label{fig:fid_fid_xl}
    \end{subfigure}
    \vspace{-0.7cm}
    \caption{\textbf{\textit{Scaling curves of FID score.}} We observe consistent performance gains and smooth scaling curves when replacing SiT's standard ViT by ViT-5. The gains are transferable across multiple model sizes and exhibit a promising trend when training longer.}
    \label{fig:fid_curve}
    % \vspace{-0.2cm}
\end{figure*}

\subsection{Image Generation}
We evaluate the transferability of ViT-5 by training it as the backbone of a Diffusion Transformer~\cite{dit} for image generation. We follow the same configuration as SiT, replacing the vanilla ViT backbone with ViT-5, and conduct training and evaluation on ImageNet-256~\cite{imagenet}. Implementation details are provided in the Appendix. As shown in Table~\ref{tab:generation}, under identical training configurations, simply replacing the original ViT backbone with ViT-5 leads to significant performance improvements over both DiT and SiT across multiple model sizes. In Table~\ref{tab:generation_7m}, we further scale training to 7M steps and observe that ViT-5 consistently outperforms its ViT counterparts in terms of Fréchet Inception Distance (FID), Inception Score (IS), as well as Precision and Recall metrics. Figure~\ref{fig:fid_curve} presents the scaling curves of diffusion models with different model sizes across varying training lengths. We observe that models using ViT-5 as the backbone consistently outperform their vanilla ViT counterparts, and exhibit smooth and stable scaling behavior as training progresses. Taken together, these results demonstrate that ViT-5 generalizes effectively across both visual understanding and image generation tasks, highlighting the broader impact of our component-wise architectural modernization. By systematically refining core Transformer components, ViT-5 serves as a strong and versatile backbone for diverse vision workloads.

\begin{table}[t]
    \centering
    \tablestyle{7pt}{1.2}
    \begin{tabular}{lrrr}
         Model & \#Params & Training steps & FID ($\downarrow$) \\
         \shline
         \dotfilled{fengblue} \textit{DiT with ViT-S} & 33M & 400k & 68.4 \\
         \dotopen{orange} \textit{SiT with ViT-S} & 33M & 400k & 57.6 \\
         \rowcolor{gray!15}
         \dotfilled{orange} \textit{SiT with ViT-5-S} & 33M & 400k & \bf 47.7 \\
         \hline
         \dotfilled{fengblue} \textit{DiT with ViT-B} & 130M & 400k & 43.5 \\
         \dotopen{orange} \textit{SiT with ViT-B} & 130M & 400k & 33.0 \\
         \rowcolor{gray!15}
         \dotfilled{orange} \textit{SiT with ViT-5-B} & 130M & 400k & \bf 29.0 \\
         \hline
         \dotfilled{fengblue} \textit{DiT with ViT-L} & 458M & 800k & 17.1 \\
         \dotopen{orange} \textit{SiT with ViT-L} & 458M & 800k & 14.9 \\
         \rowcolor{gray!15}
         \dotfilled{orange} \textit{SiT with ViT-5-L} & 458M & 800k & \bf 13.7 \\
         \hline
         \dotfilled{fengblue} \textit{DiT with ViT-XL} & 675M & 800k & 14.1 \\
         \dotopen{orange} \textit{SiT with ViT-XL} & 675M & 800k & 12.6 \\
         \rowcolor{gray!15}
         \dotfilled{orange} \textit{SiT with ViT-5-XL} & 675M & 800k & \bf 12.0 \\
    \end{tabular}
    \caption{\textbf{\textit{Class-conditional image generation}} results  on Imagenet-256. ViT-5 consistently achieves better FID over vanilla ViT with similar parameter counts. Patch size is set to 2 for all models.}
    \label{tab:generation}
    % \vspace{-0.3cm}
\end{table}

\begin{table}[t]
    \centering
    \tablestyle{6pt}{1.2}
    \begin{tabular}{lrrrrr}
         Model & FID $\downarrow$ & IS $\uparrow$ & Prec. $\uparrow$ & Recall $\uparrow$ \\
         \shline
         \dotfilled{fengblue} \textit{DiT w. ViT-XL} & 2.27 & 278.24 & 0.83 & 0.57 \\
         \dotopen{orange} \textit{SiT w. ViT-XL} & 2.06 & 277.50 & 0.83 & 0.59 \\
         \rowcolor{gray!15}
         \dotfilled{orange} \textit{SiT w. ViT-5-XL} & \textbf{1.84} & \textbf{282.73} & 0.83 & \textbf{0.60} \\
    \end{tabular}
    \caption{\textit{\textbf{Image generation with longer training.}} We train the models for 7M steps with classifier-free guidance set to 1.5. ViT-5-XL attains best performance in terms of all the listed metrics.}
    \label{tab:generation_7m}
    \vspace{-0.3cm}
\end{table}

\subsection{Dense Prediction}

We further evaluate ViT-5 on ADE20k~\cite{ade20k} for semantic segmentation using the UperNet~\cite{upernet} framework. All models are trained at a resolution of 512×512 for 160k iterations, with backbones pretrained on ImageNet-1k for the same number of epochs. All other training settings are kept identical across methods and the technical details can be found in the Appendix. As shown in Table~\ref{tab:seg}, ViT-5 consistently outperforms DeiT-III across all model scales. Specifically, ViT-5-Small, ViT-5-Base, and ViT-5-Large achieve 47.5\%, 49.1\%, and 52.0\% mIoU, respectively, compared to 45.2\%, 48.0\%, and 49.3\% for their DeiT-III counterparts under the same parameter budgets. Notably, the performance gap widens with model scale, indicating that the benefits of component-wise modernization become more pronounced in larger models. These results also demonstrate that ViT-5 transfers effectively to dense prediction tasks and delivers consistent gains without introducing task-specific architectural changes, further supporting its role as a strong and general-purpose vision backbone.

\begin{table}[t]
    \centering
    \tablestyle{6pt}{1.2}
    \begin{tabular}{lrrr}
          Backbone & Segent. head & \#Parameters & mIoU (\%) \\
          \shline
          \dotopen{orange} \textit{DeiT-III-Small} & UperNet & 42M & 45.2 \\
          \dotopen{orange} \textit{DeiT-III-Base} & UperNet & 128M & 48.0 \\
          \dotopen{orange} \textit{DeiT-III-Large} & UperNet & 354M & 49.3 \\
          \rowcolor{gray!15}
          \dotfilled{orange} \textit{ViT-5-Small} & UperNet & 42M & 47.5 \\
          \rowcolor{gray!15}
          \dotfilled{orange} \textit{ViT-5-Base} & UperNet & 128M & 49.1 \\
          \rowcolor{gray!15}
          \dotfilled{orange} \textit{ViT-5-Large} & UperNet & 354M & 52.0 \\
    \end{tabular}
    \caption{\textbf{\textit{ADE20k semantic segmentation results.}} We train the models at 512$\times$512 resolution for 160k iterations. Backbones are pretrained on ImageNet-1k for the same number of epochs. As shown, ViT-5 models consistently attain the best performance.}
    \label{tab:seg}
    \vspace{-0.5cm}
\end{table}

\begin{table*}[t]
    \centering
    \tablestyle{5pt}{1.2}
    \begin{tabular}{l|cccccccr}
         Configuration source & LayerScale & RMSNorm & SwiGLU & RoPE & Registers & QK-Norm & QKV-Bias & Acc. (\%) \\\shline
         \multicolumn{9}{l}{\textit{Designing choices taken from vision transformer models:}} \\
         Vanilla~\cite{vit} & \bxmark & \bxmark & \bxmark & \bxmark & \bxmark & \bxmark & \bcmark & 84.85 \\
         DeiT-III~\cite{deit3} & \bcmark & \bxmark & \bxmark & \bxmark & \bxmark & \bxmark & \bcmark & 85.39 \\
         DINO v2~\cite{dinov2} & \bcmark & \bxmark & \bxmark & \bxmark & \bcmark & \bxmark & \bcmark & 85.42 \\
         VisionLlama~\cite{visionllama} & \bxmark & \bxmark & \bcmark & \bcmark & \bxmark & \bxmark & \bcmark & 85.63 \\
         DINO v3~\cite{dinov3} & \bcmark & \bxmark & \bxmark & \bxmark & \bcmark & \bxmark & \bxmark & 85.50 \\
         NEPA~\cite{nepa} & \bcmark & \bxmark & \bcmark & \bcmark & \bxmark & \bcmark & \bcmark & 85.69 \\
         \hline
         \multicolumn{9}{l}{\textit{Designing choices taken from language transformer models:}} \\
         LLaMA~\cite{llama} & \bxmark & \bcmark & \bcmark & \bcmark & \bxmark & \bxmark & \bxmark & 85.73 \\
         Qwen~\cite{qwen} & \bxmark & \bcmark & \bcmark & \bcmark & \bxmark & \bxmark & \bcmark & 85.66 \\
         GPT-oss$^*$~\cite{gptoss} & \bcmark & \bcmark & \bcmark & \bcmark & \bxmark & \bxmark & \bxmark & 85.64 \\
         Gemma3~\cite{gemma3} & \bxmark & \bcmark & \bcmark & \bcmark & \bxmark & \bcmark & \bxmark & 85.75 \\
         \hline\rowcolor{gray!15}
         ViT-5 & \bcmark & \bcmark & \bxmark & \bcmark & \bcmark & \bcmark & \bxmark & \bf 86.00 \\

    \end{tabular}
    \caption{\textbf{\textit{Existing design choices are under-optimized for ViTs.}} We compare ViT-5 against the setup from prevalent vision models such as DeiT-III and DINO v2/v3. We also try transferring the configurations adapted from language models such as the LLaMA, Qwen, and Gemma series into ViTs (with RoPE extended to 2D). However, all these cases exhibit a clear performance gap compared with ViT-5. * indicates models that use post-norm. instead of explicit LayerScale. Acc. is ImageNet-1k top-1 accuracy for ViT-Large at 384 resolution.}
    \label{tab:abl}
    \vspace{-0.2cm}
\end{table*}

\subsection{Ablation Study}
\label{sec:abl}

In this work, we examine seven architectural components in ViTs and their corresponding design choices. As exhaustively enumerating all possible combinations would be prohibitively expensive, we focus on two complementary ablation settings. First, we compare ViT-5 with prevalent Transformer configurations that are commonly adopted in existing vision models or language models, in order to assess whether the design choices explored and widely used in prior work are optimal for vision tasks. Second, starting from the complete ViT-5 model, we individually ablate each of the seven components to isolate and quantify their respective contributions to overall performance.

\paragraph{Comparison to existing design.} Table~\ref{tab:abl} compares ViT-5 with architectural configurations derived from prevalent Vision Transformers and modern language models. To fully assess model capacity at larger scales and higher resolutions, this ablation is conducted using ViT-Large at a resolution of 384$\times$384. As summarized, among vision-oriented designs, models such as DeiT-III, DINO v2/v3, VisionLLaMA incorporate subsets of modern Transformer components, yet none of them simultaneously adopt all effective refinements identified in this work. As a result, their performance remains consistently below that of ViT-5, with gaps ranging from 0.31\% to over 1.15\% top-1 accuracy on ImageNet-1k.

We also observe that directly transferring configurations from language models to vision tasks is insufficient. Although LLaMA-, Qwen-, and Gemma-style setups introduce advanced normalization, gated MLPs, and relative positional encoding, they still underperform ViT-5 when adapted to ViTs (with RoPE extended to 2D). This highlights that architectural choices optimized for language models do not trivially translate to optimal vision performance. In contrast, ViT-5 achieves the best accuracy by systematically integrating these components with vision-specific considerations, demonstrating that existing design choices are under-optimized for Vision Transformers and motivating a principled, component-wise modernization.

\paragraph{Impact of individual components.} Table~\ref{tab:abl_component} reports the results of single-component ablations on ViT-5 across different model sizes. Overall, removing any individual component leads to a consistent drop in accuracy compared to the complete ViT-5 configuration, confirming that each design choice contributes positively to the final performance. While the magnitude of degradation varies across components and model sizes, no single modification dominates across all settings, underscoring the complementary nature of the adopted architectural refinements.

We also observe the impact of individual architectural components is not uniform across model scales. Certain components exhibit a stronger influence on smaller models, while others become increasingly critical as model size grows. For example, replacing GeLU with SwiGLU without proper stabilization leads to a pronounced performance drop in the Small model, reflecting the higher sensitivity of compact models to gating-induced sparsity. In contrast, components such as LayerScale and 2D RoPE show more substantial effects on larger models, where deeper networks and higher representational capacity amplify the importance of training stability and relative positional modeling. Notably, removing Registers and QK-Norm results in relatively modest degradation for Small models, but leads to consistent and larger drops for Base and Large variants, indicating that these components play a more significant role as model capacity increases.

\begin{table}[t]
    \centering
    \tablestyle{6pt}{1.2}
    \begin{tabular}{lcccc}
        Configuration & Small & Base & Large & $\Delta$ \\
        \shline
        \textit{remove LayerScale} & 82.01 & 83.86 & 84.41 & \textcolor{gray}{-0.29} \\
        \textit{RMSNorm $\rightarrow$ LayerNorm} & 82.07 & 84.00 & 84.66 & \textcolor{gray}{-0.15} \\
        \textit{GeLU $\rightarrow$ SwiGLU}$^*$ & 81.87 & 83.70 & 84.35 & \textcolor{gray}{-0.42} \\
        \textit{remove 2D RoPE} & 81.82 & 83.95 & 84.70 & \textcolor{gray}{-0.24} \\
        \textit{remove Registers} & 82.04 & 84.02 & 84.61 & \textcolor{gray}{-0.17} \\
        \textit{remove QK-Norm} & 81.99 & 84.05 & 84.78 & \textcolor{gray}{-0.12} \\
        \textit{keep QKV-bias} & 82.10 & 84.12 & 84.80 & \textcolor{gray}{-0.06} \\
        \rowcolor{gray!15}
        \textit{complete ViT-5} & 82.16 & 84.16 & 84.86 & \textcolor{gray}{N/A} \\
    \end{tabular}
    \caption{\textbf{\textit{Ablation study of single-component change.}} We report ImageNet-1k top-1 accuracy for multiple model sizes. $\Delta$ denotes the average accuracy difference relative to the complete ViT-5 model. $*$ indicates the model is affected by the over-gating issue.}
    \label{tab:abl_component}
\end{table}

\begin{table}[t]
    \centering
    \small
    \begin{tabular}{ccc}
        \begin{minipage}{0.3\linewidth}
        \centering
        \tablestyle{5pt}{1.2}
        \begin{tabular}{cc}
            Init. value & Acc. \\\shline
            1e-6 & 84.01 \\
            1e-5 & 84.05 \\\rowcolor{gray!15}
            1e-4 & \textbf{84.16} \\
        \end{tabular}
        \\[2pt]
        (a) LayerScale
        \end{minipage}
        &
        \begin{minipage}{0.3\linewidth}
        \centering
        \tablestyle{5pt}{1.2}
        \begin{tabular}{cc}
            Freq. base & Acc. \\
            \shline
            1e-5 & 84.07 \\\rowcolor{gray!15}
            1e-4 & \textbf{84.16} \\\rowcolor{white}
            1e-3 & 84.10 \\
        \end{tabular}
        \\[2pt]
        (b) RoPE
        \end{minipage}
        &
        \begin{minipage}{0.3\linewidth}
        \centering
        \tablestyle{5pt}{1.2}
        \begin{tabular}{cc}
        \#Reg. & Acc. \\
        \shline
        64 & 84.12 \\
        16 & 84.12 \\\rowcolor{gray!15}
        4  & \textbf{84.16} \\
        \end{tabular}
        \\[2pt]
        (c) Registers
        \end{minipage}
    \end{tabular}
    \caption{\textbf{\textit{Ablation of architectural hyperparameters.}} We vary the LayerScale initialization value, the frequency base of 2D RoPE, and the number of learnable registers, and report ImageNet-1k top-1 accuracy (\%) of ViT-5-B. Default setups are highlighted.}
    \label{tab:abl_detail}
    \vspace{-0.5cm}
\end{table}

\paragraph{Component details.} Table~\ref{tab:abl_detail} analyzes the sensitivity of ViT-5 to several key architectural hyperparameters. We observe that LayerScale initialization has a modest impact on performance, with too small initialization values degrades the accuracy by 0.15\% for a base-sized ViT-5 model. ViT-5 by default follows the practice of most LLMs's 1D RoPE to set a frequency base to 1e-5, which here showcases the best performance. In all ViT-5 models in this work, we employ four learnable registers. We observe that the number of registers only has trivial impact on predictive performance and shows good robustness to different setups.

\section{Conclusion}

In this paper, we conduct a systematic modernization of Vision Transformers and show that a careful component-wise refresh can unlock substantial headroom without redesigning the overall architecture. By updating normalization and activation choices, strengthening positional encoding for spatial reasoning, and introducing lightweight gating and learnable tokens, ViT-5 becomes a practical next-generation ViT that is easier to optimize and more robust across visual settings. Extensive experiments verify consistent improvements on image classification, generation and semantic segmentation. These gains transfer across tasks, indicating that the benefits are not benchmark-specific but reflect improved representations and spatial inductive biases. More broadly, our results suggest that the evolution of ViTs can follow the same modular, best-practice-driven trajectory as modern large language model backbones, where stability and performance come from principled component choices rather than increasingly complex macro-architectures. We hope ViT-5 provides a strong, compatible foundation for mid-2020s vision and multimodal systems, and encourages the community to treat backbone modernization as a first-class, reproducible design process.

\section*{Impact Statement}
The goal of this work is to advance the state of the art in Machine Learning methodologies. Given the foundational nature of our contributions, we believe the potential societal consequences align with the general progress of the field and do not present unique risks requiring specific highlighting.

\bibliography{main}
\bibliographystyle{icml2026}

\newpage
\appendix
\onecolumn
\section{Appendix}

\subsection{Image generation examples}

\begin{figure}[h]
    \centering
    \includegraphics[width=0.9\linewidth]{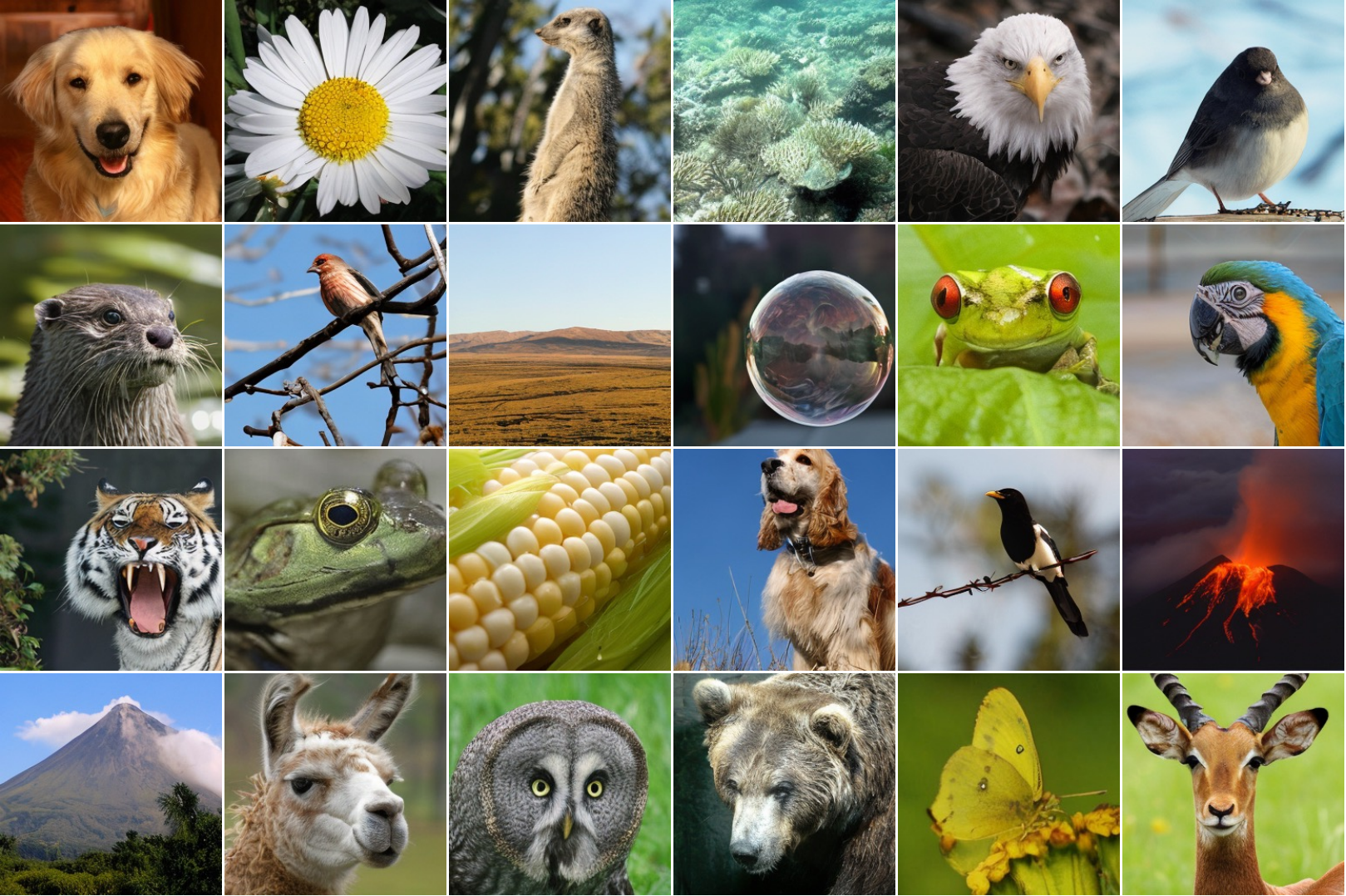}
    \caption{Image generation examples of SiT with ViT-5-XL backbone.}
    \label{fig:generated}
\end{figure}

Figure~\ref{fig:generated} presents qualitative examples generated by our model. Visually, the generated images exhibit high perceptual quality, with coherent global structure, well-preserved object shapes, and fine-grained texture details. The results appear visually pleasing and natural, without obvious artifacts or spatial inconsistencies. We attribute this strong visual quality not only to the improved representational capacity of the model, but also to its enhanced spatial awareness. Better spatial modeling allows the model to capture long-range dependencies and local details more accurately, leading to more consistent layouts and sharper textures. Together, these properties enable the model to generate images with richer details and improved visual fidelity.

\subsection{Architecture Design Principles}

Our design is guided by a set of high-level principles aimed at modernizing Vision Transformers through component-wise upgrades, rather than introducing new architectural paradigms. The primary objective is to improve representational capacity while preserving the strong generalizability of ViTs. The detailed principles include the following aspects: First, \textbf{\textit{we retain the original plain ViT structure}} to ensure broad applicability across tasks and ease of integration with multimodal systems. To this end, we avoid any form of spatial downsampling within the network. Except for the initial patchification step, all intermediate feature maps maintain a fixed spatial resolution throughout the model. Second, \textbf{\textit{we use self-attention as the only token-mixing mechanism}}. While prior work has shown that incorporating convolutions can introduce stronger inductive biases and improve performance on standard benchmarks~\cite{coatnet,maxvit,levit}, we intentionally forgo these potential gains in favor of architectural simplicity and general-purpose modeling. Third, \textbf{\textit{we do not focus on improving the patchification layer}}, and instead employ the standard non-overlapping patch embedding with linear projection. This choice is motivated by the observation that many modern generation-oriented vision models no longer rely on patch embeddings for tokenization, but instead use dedicated VAEs or learned tokenizers~\cite{dit,var}. As a result, refinements to patch embedding primarily benefit pixel-space understanding tasks and do not align with our goal of maximizing generalizability.

\subsection{More attention visualization}
\begin{figure}[h]
    \centering
    \includegraphics[width=0.8\linewidth]{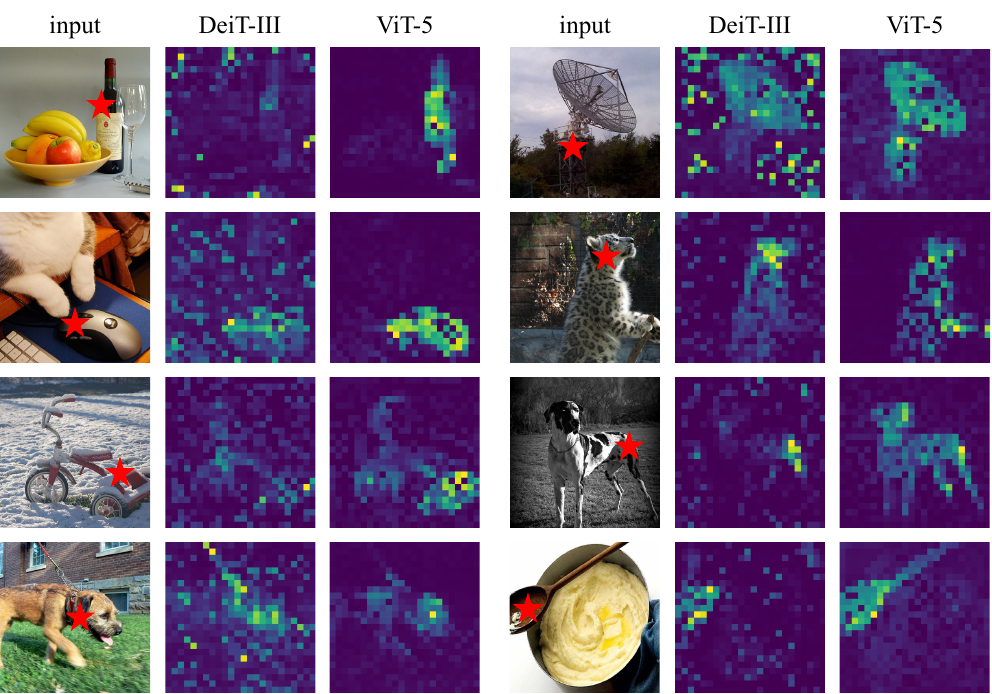}
    \caption{Attention visualization of local tokens.}
    \label{fig:attn_local}
\end{figure}

Figure~\ref{fig:attn_local} visualizes the attention maps of local tokens for DeiT-III and ViT-5 under the same experimental setting as Figure~\ref{fig:attn}, except that we attend from a selected local token instead of the class token. Both models are pretrained on ImageNet-1K for the same duration and fine-tuned at a resolution of 384×384; attention visualizations are generated at the same resolution. Compared to DeiT-III, ViT-5 exhibits noticeably cleaner and more structured attention patterns. The attention of local tokens in ViT-5 is more spatially coherent and better aligned with semantically meaningful regions, while background artifacts are largely suppressed. In contrast, DeiT-III often shows more diffuse and noisy activations. Together with the class-token attention visualizations in the main text, these results qualitatively demonstrate that ViT-5 achieves a substantive improvement in spatial modeling capability, beyond numerical performance gains. The cleaner attention maps indicate more precise token interactions and more reliable spatial reasoning, highlighting the effectiveness of the component-wise modernization introduced in ViT-5.

\subsection{Technical Details}

\paragraph{ImageNet classification.} For ImageNet-1k classification, ViT-5 follows a two-stage training pipeline, consisting of large-scale pretraining and subsequent fine-tuning. Overall, our training recipe largely follows that of DeiT-III, with several minor modifications to accommodate the architectural changes introduced in ViT-5. In the pretraining stage (Table~\ref{tab:vit5_pt}), models are trained from scratch using the LAMB optimizer~\cite{lamb} with a large batch size and cosine learning rate schedule. We adopt standard data augmentations commonly used in DeiT-style training, including random resized cropping, horizontal flipping, Mixup, and CutMix, while disabling label smoothing and dropout. Stochastic depth is applied with scale-dependent rates, and gradient clipping is enabled for training stability. In the fine-tuning stage (Table~\ref{tab:vit5_ft}), we switch to the AdamW optimizer with a smaller learning rate and train for a short schedule. Fine-tuning is performed at resolutions of 224 and 384 when applicable, following common practice. Compared to pretraining, fine-tuning uses a simplified augmentation strategy with label smoothing enabled and adjusted stochastic depth rates. These settings ensure a fair and strong training baseline, while allowing the effects of ViT5’s architectural refinements to be evaluated independently of aggressive recipe changes.

\paragraph{Image generation.} For image generation, we follow the training protocol of SiT~\cite{sit} and replace the vanilla ViT backbone with ViT-5. Unless otherwise specified, all hyper-parameters and optimization settings are kept identical to those in SiT to ensure a fair comparison. In particular, models are trained using the same optimizer, batch size, learning rate schedule, and data preprocessing as described in the original SiT paper. We consider two training regimes. First, we perform short training runs with a maximum of 800K steps across different model sizes to evaluate efficiency and early-stage scaling behavior. Second, to assess long-horizon scalability, we conduct a large-scale training run of ViT-5-XL for 7M steps. This long training experiment adopts a classifier-free guidance scale of 1.5 and uses the stochastic differential equation (SDE) formulation, following the SiT setup.

\paragraph{Semantic segmentation.} For semantic segmentation, we evaluate ViT-5 on ADE20K using the UperNet framework. Our training protocol largely follows prior ViT-based segmentation works, and is kept identical across backbones to ensure fair comparison. All models are initialized from ImageNet-1k pretrained checkpoints and trained at a resolution of 512×512 for 160k iterations. We use standard data augmentation strategies for dense prediction, including random resizing, cropping, and horizontal flipping. Optimization is performed using AdamW with a cosine learning rate schedule, and stochastic depth is applied in a scale-dependent manner. No task-specific architectural modifications are introduced when replacing the vanilla ViT backbone with ViT-5.

\begin{table}[h]
    \centering
    \tablestyle{6pt}{1.2}
    \begin{tabular}{l|ccc}
        configuration & ViT-5-small & ViT-5-base & ViT-5-large \\
        \shline
        batch size &  & 2048 &  \\
        optimizer &  \multicolumn{3}{c}{LAMB~\cite{lamb}} \\
        loss &  & BCE &  \\
        lr & 4e-3 & 3e-3 & 3e-3 \\
        lr decay &  & cosine &  \\
        weight decay &  & 0.05 &  \\
        warmup epochs &  & 5 &  \\
        input size & 224$\times$224 & 192$\times$192 & 192$\times$192 \\
        epochs & 800 & 800 & 400 \\
        \hline
        label smoothing $\epsilon$ &  & \xmark &  \\
        dropout &  & \xmark &  \\
        stochastic depth & 0.05 & 0.2 & 0.35 \\
        repeated augmentation &  & \cmark &  \\
        gradient clipping &  & 1.0 &  \\
        \hline
        horizontal flip &  & \cmark &  \\
        random resized crop &  & \cmark &  \\
        rand augment &  & \xmark &  \\
        3 augment~\cite{deit3} &  & \cmark &  \\
        mixup alpha &  & 0.8 &  \\
        cutmix alpha &  & 1.0 &  \\
        random erasing prob. &  & \xmark &  \\
        color jitter &  & 0.3 &  \\
        test crop ratio &  & 1.0 &  \\
    \end{tabular}
    \caption{ViT-5 pretraining hyper-parameters on ImageNet-1k.}
    \label{tab:vit5_pt}
\end{table}

\begin{table}[t]
    \centering
    \tablestyle{6pt}{1.2}
    \begin{tabular}{l|ccc}
        configuration & ViT-5-small & ViT-5-base & ViT-5-large \\
        \shline
        batch size &  & 512 &  \\
        optimizer & & AdamW & \\
        loss &  & CE &  \\
        lr &  & 1e-5 & \\
        lr decay &  & cosine &  \\
        weight decay &  & 0.1 &  \\
        warmup epochs &  & 5 &  \\
        input size & 224 & 224 and 384 & 224 and 384 \\
        epochs & & 20 & \\
        \hline
        label smoothing $\epsilon$ &  & 0.1 &  \\
        dropout &  & \xmark &  \\
        stochastic depth & 0.05 & 0.25 & 0.5 \\
        repeated augmentation &  & \xmark &  \\
        gradient clipping &  & \xmark &  \\
        \hline
        horizontal flip &  & \cmark &  \\
        random resized crop &  & \cmark &  \\
        rand augment &  & \cmark &  \\
        3 augment~\cite{deit3} &  & \xmark &  \\
        mixup alpha &  & 0.8 &  \\
        cutmix alpha &  & 1.0 &  \\
        random erasing prob. &  & 0.25 &  \\
        color jitter &  & 0.3 &  \\
        test crop ratio &  & 1.0 &  \\
    \end{tabular}
    \caption{ViT-5 finetuning hyper-parameters on ImageNet-1k.}
    \label{tab:vit5_ft}
\end{table}

\end{document}